\documentclass{article}
\usepackage[T1]{fontenc} 
\usepackage[utf8]{inputenc} 
\usepackage{spconf,amsmath,cite,url}
\usepackage{graphicx}
\usepackage{color}

\title{Multi-Step Chord Sequence Prediction Based on\\Aggregated Multi-Scale Encoder-Decoder Networks}

\begin{document}

\setlength{\abovedisplayskip}{6pt}
\setlength{\belowdisplayskip}{6pt}
\allowdisplaybreaks[4]
\setlength\floatsep{5pt}
\setlength\textfloatsep{10pt}
\setlength\intextsep{5pt}
\setlength\abovecaptionskip{0pt}

\maketitle
\begin{abstract}
 This paper studies the prediction of chord progressions for jazz music
 by relying on machine learning models. The motivation of our study comes from the recent success of neural networks for performing automatic music composition. Although high accuracies are obtained in single-step prediction scenarios, most models fail to generate accurate multi-step chord predictions. 
In this paper, we postulate that this comes from the multi-scale structure of musical information and propose new architectures based on an iterative temporal aggregation of input labels. Specifically, the input and ground truth labels are merged into increasingly large temporal bags, on which we train a family of encoder-decoder networks for each temporal scale. In a second step, we use these pre-trained encoder bottleneck features at each scale in order to train a final encoder-decoder network.
Furthermore, we rely on different reductions of the initial chord alphabet into three adapted chord alphabets. We perform evaluations against several state-of-the-art models and show that our multi-scale architecture outperforms existing methods in terms of accuracy and perplexity, while 
requiring relatively few parameters. 
We analyze musical properties of the results, showing the influence of downbeat position within the analysis window on accuracy, and evaluate errors using a musically-informed distance metric.
\end{abstract}
\section{Introduction}\label{sec:introduction}

Most of today's Western music is based on an underlying harmonic structure. This structure describes the progression of the piece with a certain degree of abstraction and varies at the scale of the pulse. It can therefore be represented by a "chord sequence", with a chord representing the harmonic content of a beat.
Hence, real-time music improvisation system, such as \cite{nika2017dyci2}, crucially need to be able to predict chords in real time along with a human musician at a long temporal horizon. Indeed, chord progressions aim for definite goals and have the function of establishing or contradicting a tonality \cite{schoenberg1969structural}. 
A long-term horizon is thus necessary since these structures carry more than the step-by-step conformity of the music to a local harmony. Specifically, the problem can be formulated as: given a history of beat-aligned chords, output a predicted sequence of future beat-aligned chords \textit{at a long temporal horizon}. In this paper, we use a set of ground truth chord sequences as input, but the model described here could be combined with an automatic chord extractor \cite{carsault2018using, korzeniowski2016fully} for use in a complete improvisation system.

Most chord estimation systems combine a temporal model and an acoustic model, in order to estimate chord changes and timing at the audio frame level 
\cite{boulanger2013audio, korzeniowski2018improved}. 
Even if such models analyze the temporal structure of chord sequences, our task is different. Indeed, we want to predict future chords symbols without any additional acoustic information at each step of the prediction.

In this paper we use the term multi-step chord sequence generation for the prediction of a series of possible continuing chords according to an input sequence.
Most existing systems for multi-step chord sequence generation only target the prediction of the next chord symbol given a sequence, disregarding repeated chords and ignoring timing \cite{tsushima2018generative, scholz2009robust}. Exact timing is important for our use case, and such models cannot be used without retraining them on sequences including repeated chords. However, since the "harmonic rhythm" (frequency at which the harmony changes) is often 2, 4, or even 8 beats in the music of our study area, such models inherently cannot generalize to real-life scenarios, and can be outperformed by a simple identity function \cite{crestel2017live}. Moreover, such predictive models can suffer from error propagation if used to predict more than a single chord at a time. Since we want to use our chord predictor in a real-time improvisation system \cite{nika2017dyci2, nika2016guiding}, the ability to predict coherent long-term sequences is of utmost importance.

\begin{figure}
\centering
\includegraphics[width=0.45\textwidth]{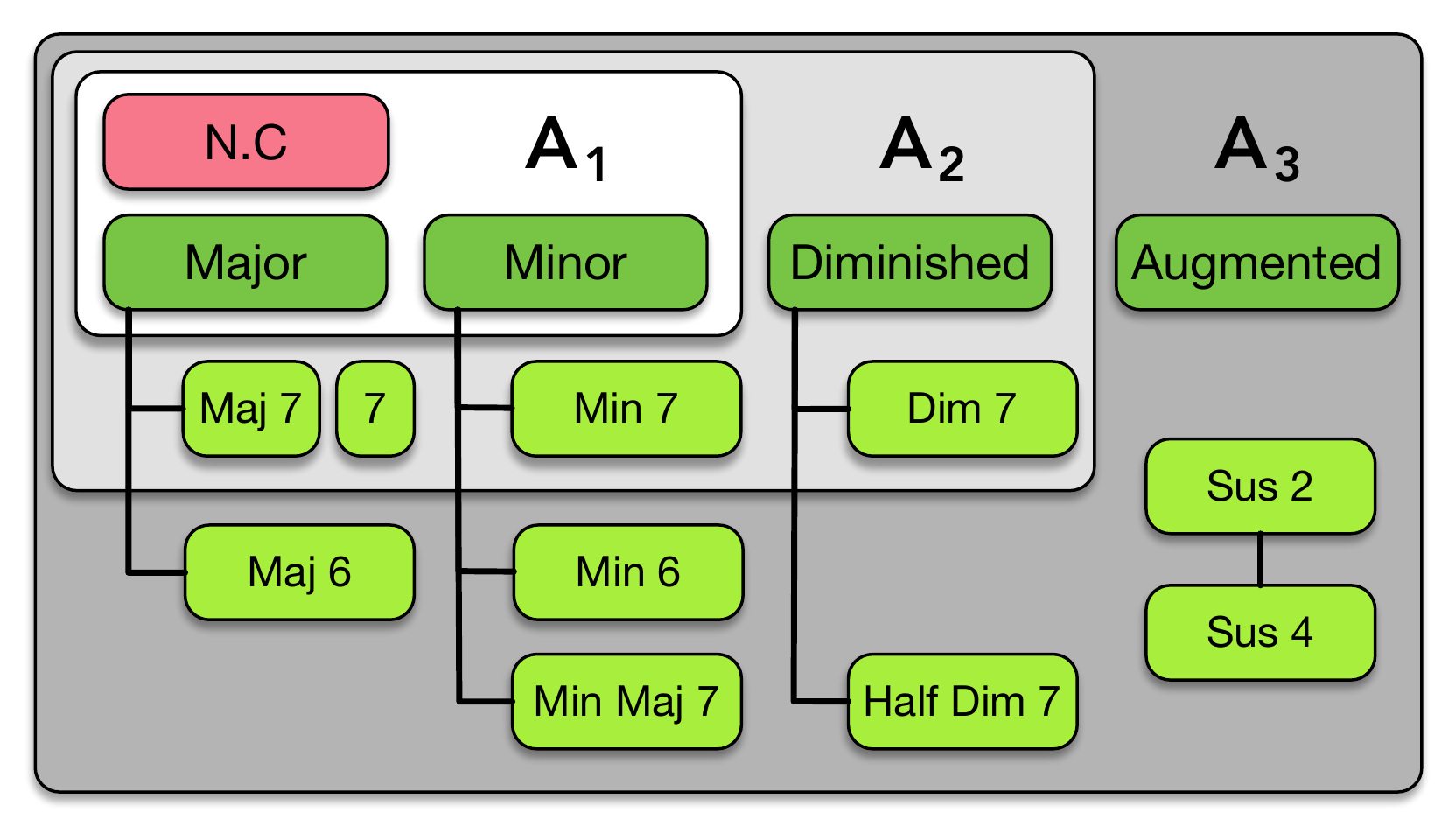}
\caption{\label{fig:vocab} The three chord vocabularies $A_1$, $A_2$, and $A_3$ we use in this paper are defined as increasingly complex sets. The standard triads are shown in dark green.}
\end{figure}

In this paper, we study the prediction of a sequence of 8 beat-aligned chords given the 8 previous beat-aligned chords. The majority of chord extraction and prediction studies rely on a fixed chord alphabet of 25 elements (major and minor chords for every root note (i.e. c, c\#, d, d\#, etc.), along with a \textit{no chord} symbol), whereas some studies perform an exhaustive treatment of every unique set of notes as a different chord \cite{yoshii2011vocabulary,eigenfeldt2010realtime,paiement2005probabilistic}. Here, we investigate the effect of using chord alphabets of various precision. We use previous work \cite{carsault2018using, mcfee2017structured} to define three different chord alphabets (as depicted in Figure~\ref{fig:vocab}), and perform an evaluation using each of them.

We propose a multi-scale model which predicts the next 8 chords directly, eliminating the error propagation issue which inherently exists in single-step prediction models. In order to provide a multi-scale modeling of chord progressions at different levels of granularity, we introduce an aggregation approach, summarizing input chords at different time scales.
First, we train separate encoder-decoders to predict the aggregated chords sequences at each of those time scales. Finally, we concatenate the bottleneck layers of each of those pre-trained encoder-decoders and train the multi-scale decoder to predict the non-aggregated chord sequence from the concatenated encodings. This multi-scale design allows our model to capture the higher-level structure of chord sequences, even in the presence of multiple repeated chords.

To evaluate our system, we compare its chord prediction accuracy to a set of various state-of-the-art models. 
We also introduce a new musical evaluation process, which uses a musically informed distance metric to analyze the predicted chord sequences.

This paper is organized as follows. Section \ref{sec:prev} contains related work, and  Section \ref{sec:prop} describes our proposed model. Experimental setup is described in Section \ref{sec:exp}, and we discuss the results in Section \ref{sec:resul}. Finally, a conclusion and ideas for future work can be found in Section \ref{sec:conc}.

\section{Previous work}
\label{sec:prev}

Most works in chord sequence prediction focus on chord transitions (eliminating repeated chords), and does not include the duration of the chords. Such models include Hidden Markov Models (HMMs) and N-Gram models \cite{tsushima2018generative, scholz2009robust, yoshii2011vocabulary}. Here, we use a 9-gram model, trained at the beat level, as a baseline comparison. HMMs 
\cite{eigenfeldt2010realtime}
have also been used for chord sequence estimation based on the melody or bass line, sometimes by including a duration component. However, they rely on the underlying melody to generate an accompanying harmonic progression, rather than predicting a future chord sequence. Recently, neural models for audio chord sequence estimation have also been proposed, but these similarly rely on the underlying audio signal during estimation \cite{boulanger2013audio, korzeniowski2018improved}.

Long Short-Term Memory (LSTM) networks have shown some promising results in chord sequence generation. For instance, \cite{eck2002finding} describes an LSTM which can generate a beat-aligned chord sequence along with an associated monophonic melody. Similarly, in a recent article \cite{choi2016text}, a text-based LSTM is used to perform automatic music composition. The authors use different types of Recurrent Neural Networks (RNNs) to generate beat-aligned symbolic chord sequences. They focus on two different approaches, each with the same basic LSTM architecture: a \textit{word-RNN}, which treats each chord as a single symbol, and a \textit{char-RNN}, which treats each character in a chord's text-based transcription as a single symbol (in that case, \texttt{A:min} is a sequence of 5 symbols). In this paper, we re-implemented the same word-RNN model as a baseline for comparison. However, we aim to improve the learning by embedding specific multi-step prediction mechanisms, in order to reduce single-step error propagation.

\subsection{Multi-step prediction}

It has been observed that using an LSTM for multi-step prediction 
can suffer from error propagation, where the model is forced to re-use incorrectly predicted steps \cite{cheng2006multistep}.
Indeed, at inference time, the LSTM cannot rely on the ground-truth sequence and is forced to rely on samples from its previous output distribution.
Thus, the predicted sequences gradually diverge as the error propagates and gets amplified at each step of the prediction.
Another issue is that the dataset of chord sequences contains a large amount of repeated symbols. Hence, the easiest error minimization for networks would be to approximate the identity function, by always predicting the next symbol as repeating the previous one.
In order to mitigate this effect, previous works \cite{choi2016text} introduce a diversity parameter that re-weights the LSTM output distribution at each step in order to penalizes redundancies in the generated sequence. Instead, we propose to minimize this repetition, as well as error propagation, by feeding the LSTM non-ground truth chords during training time using teacher forcing \cite{williams1989learning} (see Section \ref{sec:lstm}).
We also propose to generate the entire sequence of chords directly using a multi-scale feed-forward model.

\section{Proposed Method}\label{sec:prop}

\begin{figure*}
\centering
\includegraphics[width=0.8\textwidth]{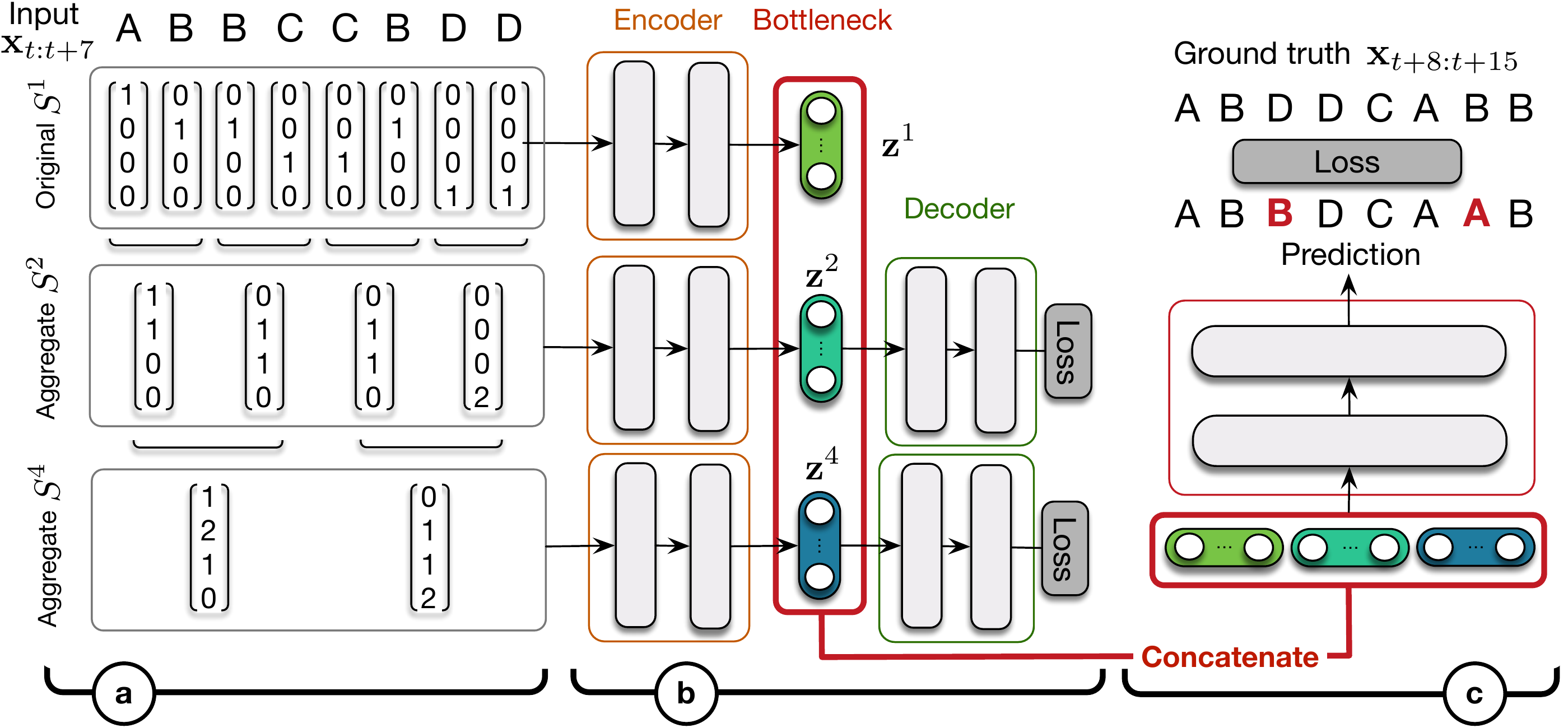}
\caption{\label{fig:MSae} The architecture of our proposed system.}
\end{figure*}

Our proposed approach is based on a sequence-to-sequence architecture, which is defined by two major components. The first, called an \textit{encoder}, takes the input sequence and transforms it into a latent vector. This vector is then used as input to the second \textit{decoder} network, which generates the output sequence. 
Our architecture is a combination of Encoder-Decoder (ED) networks  \cite{cho2014properties} and an aggregation mechanism for pre-training the models at various time scales.
Here, we define \textit{aggregation} as an increase of the temporal span covered by one point of chord information in the input/target sequences (as depicted in Figure~\ref{fig:MSae}-a).
In order to train our whole architecture, we use a two-step training procedure. During the first step, we train separately each network with inputs and targets aggregated at different ratios (Figure~\ref{fig:MSae}-b). The second step performs the training of the whole architecture where we concatenate the output of the previously trained encoders (Figure~\ref{fig:MSae}-c).

\subsection{Multi-scale aggregation}
In this work, each chord is represented by a categorical one-hot vector. We compute aggregated inputs and targets by repeatedly increasing the temporal step of each sequence by a factor of two, and computing the sum of the input vectors within each step. This results in three input/output sequence pairs: $S^1$ and $T^1$, the original one-hot sequences; $S^2$ and $T^2$, the sequences with timestep 2; and $S^4$ and $T^4$, the sequences with timestep 4. Formally, for each timestep greater than 1, $S^n_i$ (the $i$th vector of $S^n$, 0-indexed) is calculated as shown in Equation \ref{eq:agg}. This aggregation is illustrated in Figure~\ref{fig:MSae}-a.

\begin{equation}
    S_{i}^{n} = S_{2i}^{n/2} + S_{2i+1}^{n/2}
    \label{eq:agg}
\end{equation}

\subsection{Pre-training networks on aggregated inputs/targets}
First, we train two ED networks: one for each of the aggregated input/target pairs.
In order to obtain informative latent spaces we create a bottleneck between the \textit{encoder} and the \textit{decoder} networks, which forces the network to compress the input data. Hence, we first train each ED network independently with aggregated inputs and targets at different ratio. Our loss function for this training is the Mean Squared Error between $S^n$ and $T^n$. Then, from each encoder, we obtain the latent representation $z^n$ of its input sequence $S^n$.

\subsection{Second training of the whole architecture}
For the full system, we take the latent representations of the pre-trained ED networks, and concatenate them with the latent vector of a new ED network whose input is the original sequence $S^1$. From this concatenated latent vector, we train a decoder through the cross-entropy loss to the target $T^1$. During this full-system training, the parameters of the pre-trained independent encoders are frozen and we optimize only the parameters of the non-aggregated ED.




\section{Experiments}\label{sec:exp}

\subsection{Dataset}

In our experiments, we use the \textit{Realbook} dataset \cite{choi2016text}, which contains 2,846 jazz songs based on band-in-a-box files\footnote{\url{http://bhs.minor9.com/}}. All files come in a \textit{xlab} format and contain time-aligned beat and chord information. We choose to work at the beat level, by processing the xlab files in order to obtain a sequence of one chord per beat for each song. 
We perform a 5-fold cross-validation by randomly splitting the song files into training (0.6), validation (0.2), and test (0.2) sets with 5 different random seeds for the splits. We report results as the average of the resulting 5 scores. We use all chords sub-sequences of 8 elements throughout the different sets, beginning at the first No-chord symbol (padding this input, and the target being chords 2 to 9), and ending where the target is the last 8 chords of each song.

\subsection{Alphabet reduction}\label{subsec:alph}

The dataset is composed by a total alphabet of 1259 chord labels. This great diversity comes from the precision level of the chosen syntax.
Here, we apply a hierarchical reduction of the original alphabet into three smaller alphabets of varying levels of specificity as depicted in Figure~\ref{fig:vocab}, containing triads and tetrachords commonly used to write chord progressions. Each node in the figure (except N.C. for no chord) represents 12 chord symbols (one for each non-enharmonic root note). Dark green represents the four standard triads: major, minor, diminished, and augmented. $A_1$ contains 25 symbols, $A_2$ contains 85 symbols, and $A_3$ contains 169 symbols. The black lines represent chord reductions, and chord symbols not in a given alphabet are either reduced to the corresponding standard triad, 
or replaced by the no chord symbol.

\subsection{Models and training}

In order to evaluate our proposed model, we compare it to several state-of-the-art methods for chord predictions. In this section, we briefly introduce these models and the different parameters used for our experiments.

\paragraph*{Naive Baselines.} We compare our models against two naive baselines: predicting a \textit{random} chord at each step; and predicting the \textit{repetition} of the most recent chord.

\paragraph*{N-grams.}
The N-gram model estimates the probability of a chord occurring given the sequence of the previous $n-1$ chords. Here, we use $n=9$ (a 9-gram model), and train the model using the Knesser-Ney smoothing \cite{heafield2013scalable} approach. For training, we replace the padded N.C. symbols with a single start symbol. Since an n-gram model does not require a validation set, we combine this with the training set for training the n-gram. During decoding, we use a beam search, saving only the top 100 states (each of which contains a sequence of 9 chords and an associated probability) at each step. The probability of a chord at a given step is calculated as the sum of the (normalized) probabilities of the states in the beam at that step which contain that chord as their most recent chord.

\paragraph*{LSTM.}
\label{sec:lstm}

In our experiments, we use the teacher forcing algorithm 
\cite{williams1989learning} to train our LSTM. Given an input sequence and a target sequence, the free training algorithm uses the predicted output at time $t-1$ to compute predicted output at time $t$. In our training we use the ground truth data from the time $t-1$ or the the predicted output at time $t-1$ randomly to compute the predicted output at time $t$.

We use the Seq2Seq architecture to build our model \cite{sutskever2014sequence}. Thus, our network is divided into two parts (encoder and decoder). The encoder extracts useful information of the input sequences and gives this hidden representation to the decoder, which generates the output sequence. We did a grid search to find correct layer size (see Table~\ref{tab:tablarch} for details on the architecture). We add a dropout layer ($p=0.5$) between each layer and a Softmax at the output of the decoder. Our models are trained with ADAM and a learning rate of $1e^{-4}$.

\paragraph*{MLP-ED and Multi-scale ED}

We compare our model to a MLP Encoder Decoder (MLP-ED). We observed that adding a bottleneck between the encoder and the decoder slightly improved the results compared to the classical MLP.
All encoder and decoder blocks are defined as fully-connected layers with ReLU activation. A simple grid search
defined that the size of 50 hidden units was the most appropriate for the bottleneck. The architectures and parameters of all our tested models are summarized in Table~\ref{tab:tablarch}.

The Multi-Scale ED is composed of the same encoder and decoder layers as the MLP-ED in terms of their parameters.
As Table~\ref{tab:tablarch} shows, the proposed Multi-Scale AutoEncoder model has more parameters than the MLP-ED, but fewer than the LSTM. For these ED networks we add a dropout layer ($p=0.5$) between each layer and a Softmax layer at the output of our decoder. Our models are trained with ADAM Optimizer with a learning rate of $1e^{-4}$.


\begin{table}
 \begin{center}

\begin{tabular}{l|c|c|c}
  \hline
   & MLP-ED & LSTM & MultiScale-ED \\
  \hline
  $\sharp$ encoder layers & 2  & 2 & 2 \\ 
  $\sharp$ decoder layers & 2  & 2 & 2 \\
  $\sharp$ hidden units & 500  & 500 & 500 \\ 
  $\sharp$ bottleneck dims & 50  & - & 50 \\
  \hline
  $\sharp$ parameters on $A_1$ & 0.75M & 6.9M & 2.1M \\
  \hline
\end{tabular}
  \caption{Parameters for the different neural networks.}\label{tab:tablarch}
\end{center}
\end{table}

\section{Results}\label{sec:resul}

\subsection{Quantitative analysis}

We trained all models on the three alphabets described in Section~\ref{subsec:alph}. In order to evaluate our models, we compute the mean prediction accuracy over the output chord sequences (see Table~\ref{tab:tablres}). The first two lines represent the accuracy over increasingly complex alphabets for the \textit{random} and \textit{repeat} models. Interestingly, the \textit{repeat} classification score remains rather high, even for the most complex alphabets, which shows how common repeated chords are in our dataset.
The last four lines show the accuracy of the more advanced models, where we can observe that the score decreases as the alphabet becomes more complex.

\begin{table}
 \begin{center}

\begin{tabular}{l|c|c|c}
  \hline
  Model & $A_1$ & $A_2$ & $A_3$ \\
  \hline
  Random & 4.00 & 1.37 & 0.59 \\ 
  Repeat & 34.2 & 31.6 & 31.1 \\
  \hline
  \hline
  9-Gram  & 40.4 & 37.8 & \textbf{36.9} \\ 
  \hline
  MLP-ED & 41.8 & 37.0 & 35.2 \\
  LSTM & 41.8 & 37.3 & 36.0 \\
  \hline
  \hline
  MS-ED & \textbf{42.3} & \textbf{38.0} & 36.5 \\
  \hline
\end{tabular}
  \caption{Mean prediction accuracy for each method over the different chord alphabets.} \label{tab:tablres}
\end{center}
\end{table}

First, we can observe that our Multi-Scale ED obtains the highest results in most cases, outperforming the LSTM in all scenarios. However, the score obtained with a 9-Gram on $A_3$ is higher than the Multi-Scale ED. We hypothesize that this can be partly explained by the distribution of chord occurrences in our dataset. Many of the chords in $A_3$ are very rare, and the neural models may simply need more data to perform well on such chords. The 9-gram, on the other hand, uses smoothing to estimate probabilities of rare and unseen chords (though it is likely that it would not continue to improve as much as the neural models, given more data).

We also compare our models in terms of perplexity and rank of the correct target chord in the output probability vector (see Table~\ref{tab:preprank}). Our proposed model performs better or equal to all other models on all alphabets with these metrics, which are arguably more appropriate for evaluating performance on our task.








\begin{table}
 \begin{center}

\begin{tabular}{l|c|c|c||c|c|c}
  \hline
  Measure & \multicolumn{3}{c||}{Perplexity} & \multicolumn{3}{c}{Rank}\\
  \hline
  Alphabet & $A_1$ & $A_2$ & $A_3$ & $A_1$ & $A_2$ & $A_3$ \\
  \hline
  9-Gram  & 7.93 & 13.3 & \textbf{15.7} & 4.13 & 8.05 & 10.3\\ 
  MLP-ED & 7.45 & 13.5 & 16.7 & 3.98 & 8.08 & 10.6 \\
  LSTM & 7.60 & 13.3 & 16.0 & 4.02 & 7.94 & 10.2 \\
  MS-ED & \textbf{7.40} & \textbf{12.9} &  \textbf{15.7} & \textbf{3.94} & \textbf{7.74} & \textbf{9.99} \\
  \hline
\end{tabular}
  \caption{Left side: Perplexity of each model over the test dataset; Right side: Mean of the rank of the target chords in the output probability vectors.} \label{tab:preprank}
\end{center}

\end{table}

\begin{table}
 \begin{center}

\begin{tabular}{l|c|c|c||c|c|c}
  \hline
  Level & \multicolumn{3}{c||}{Probabilistic} & \multicolumn{3}{c}{Binary}\\
  \hline
  Alphabet & $A_1$ & $A_2$ & $A_3$ & $A_1$ & $A_2$ & $A_3$ \\
  \hline
  9-Gram  & 1.66 & 1.61 & 1.57 & 1.33 &  1.30 & \textbf{1.28} \\ 
  MLP-ED & 1.61 &  1.61 & 1.58 &  1.28 & 1.31 & 1.31 \\
  LSTM &  1.60 & 1.59 & \textbf{1.54} & 1.29 & 1.30 & 1.29 \\
  MS-ED & \textbf{1.59} & \textbf{1.58} &  1.55 & \textbf{1.28} & \textbf{1.29} & \textbf{1.28} \\
  \hline

\end{tabular}
    \caption{Mean Euclidean distance between (left) contribution of all the chords in the output probability vectors, and (right) chords with the highest probability score in the output vectors.} \label{tab:musicalDist}
\end{center}

\end{table}


\subsection{Musical analysis}
\subsubsection{Euclidean distance}
In order to compare different models, we evaluate errors through a musically-informed distance described in \cite{carsault2018using}. In this distance, each chord is associated with a binary pitch class vector. Then, we compute the Euclidean distance between the predicted vectors and target chords.

The results, presented in Table~\ref{tab:musicalDist}, show two different approaches of using this distance. The left side of the table represents the Euclidean distances between the contribution
of all the chords in each model's output probability vector (weighted by their probability) and the target chord vectors.
The right side shows the Euclidean distance between the single most likely predicted chord at each step and the target chord. We observe that the Multi-Scale ED always obtains the best results, except on a single case ($A_3$ and probabilistic distance), where the LSTM performs best by a small margin.

\subsubsection{Influence of the downbeat position in the sequence}

\begin{figure}
\centering
\includegraphics[width=0.49\textwidth]{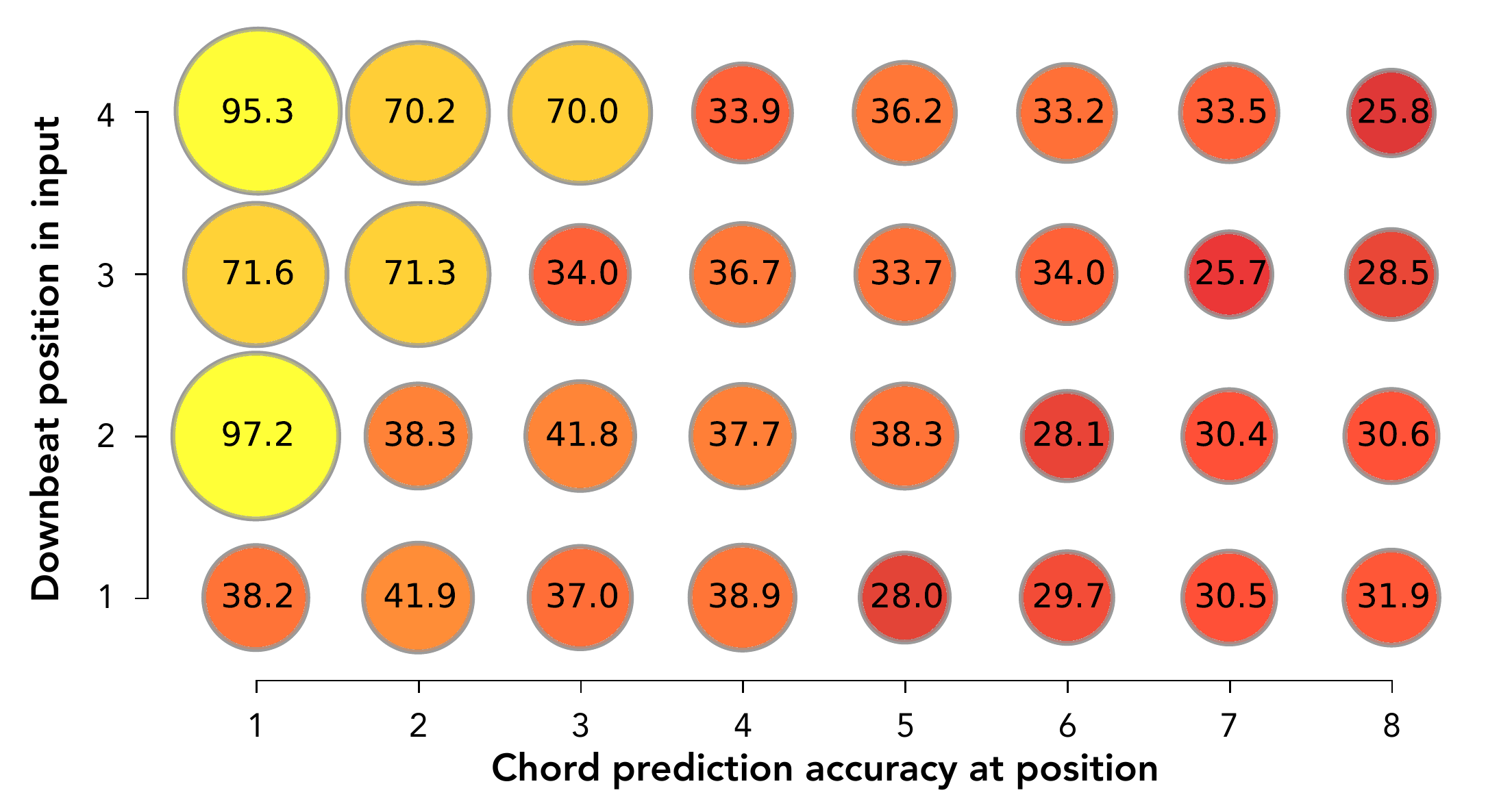}
\caption{\label{fig:accBeat} Chord prediction accuracy at each position of the output sequence depending on the position of the downbeat in the input sequence. 
}
\end{figure}

Figure~\ref{fig:accBeat} shows the prediction accuracy of the Multi-Scale ED on $A_1$ at each position of the predicted sequence, depending on the position of the downbeat in the input sequence. 
Prediction accuracy significantly decreases across each bar line, likely due to bar-length repetition of chords. 
The improvement of the score for the first position when the downbeat is in position 2 can certainly be explained by the fact that the majority of the RealBook tracks have a binary metric (often 4/4). 
We also see that the prediction accuracy of chords on downbeats is lower than that of the following chords in the same bar. 
It can be assumed that this is due to the fact that chords often change on the downbeat, and that the following target chords can sometimes have the same harmonic function as the predicted chords but without being exactly the same. This approach could be studied using a more functional approach to harmony as presented in \cite{carsault2018using}.
Both trends are observed over all models and alphabets. This underlines the importance of using downbeat position information in the development of future chord sequence prediction models.

\section{Conclusion}
\label{sec:conc}

In the paper, we studied the prediction of beat-synchronous chord sequences at a long horizon. We introduced a novel architecture based on the aggregation of multi-scale encoder-decoder networks. We evaluated our model in terms of accuracy, perplexity and rank over the predicted sequence, as well as by relying on musically-informed distances between predicted and target chords.

We showed that our proposed approach provides the best results for simpler chord alphabets in term of accuracy, perplexity, rank and musical evaluations. For the most complex alphabet, existing methods appear to be competitive with our approach and should be considered. Our experiments on the influence of the downbeat position in the input sequence underlines the complexity of predicting chords across bar lines. For future work, we intend to investigate the use of the downbeat position in chord sequence prediction systems.

\section{Acknowledgments}
This work was supported by the MAKIMOno project 17-CE38-0015-01  funded  by  the  French  ANR  and  the  Canadian  NSERC(STPG 507004-17), the ACTOR Partnership funded by the Canadian SSHRC (895-2018-1023) and an NVIDIA GPU Grant. 
It was also supported in part by JST ACCEL No.~JPMJAC1602, JSPS KAKENHI No.~16H01744 and No.~19H04137, and the Kyoto University Foundation.

\bibliographystyle{IEEEtran}
\bibliography{main}

%
%
%
%

\end{document}